\documentclass[12pt]{article}

\usepackage[T1]{fontenc}
\usepackage[utf8]{inputenc}
\usepackage{lmodern}           

\usepackage[a4paper, margin=2.5cm]{geometry}
\usepackage{graphicx}
\usepackage{booktabs}
\usepackage{hyperref}
\usepackage{subcaption}
\usepackage[font=small]{caption}
\usepackage{placeins}

\usepackage{amsmath}
\usepackage{amssymb}

\usepackage{makecell}

\usepackage{float}

\usepackage[authoryear]{natbib}
\usepackage{hyperref}

\usepackage{tipa}
\usepackage{tipx}              

\newcommand{\ipa}[1]{\textipa{#1}}

\title{\textbf{The Dynamic Articulatory Model DYNARTmo: Dynamic Movement Generation and Speech Gestures}}

\author{
  Bernd J.~Kröger$^{1,2}$\\[1ex]
  {\small $^{1}$Medical School, RWTH Aachen University, Aachen, Germany}\\
  {\small $^{2}$Kröger Lab, Belgium, \url{www.speechtrainer.eu}}\\[2ex]
}

\date{}

\begin{document}
\maketitle

\begin{abstract}
This paper describes the current implementation of the dynamic articulatory model \textbf{DYNARTmo}, 
which generates continuous articulator movements based on the concept of \textit{speech gestures} 
and a corresponding \textit{gesture score}. The model provides a neurobiologically inspired 
computational framework for simulating the hierarchical control of speech production—from linguistic 
representation to articulatory-acoustic realization. We present the structure of the gesture inventory, 
the coordination of gestures in the gesture score, and their translation into continuous articulator 
trajectories controlling the DYNARTmo vocal tract model.
\end{abstract}

\section{Introduction}

The neural generation and control of speech utterances is a complex process that is still not fully 
understood. However, several \textit{neurobiologically inspired models} have been proposed that 
describe the hierarchical control concept of utterance generation (e.g., \citet{Hickok_Poeppel_2012}; 
\citet{Bohland_Bullock_Guenther_2010}; \citet{Kroeger_etal_2020}; \citet{Parrell_etal_2018}). 
This process begins with the neural activation of the \textit{cognitive-linguistic representation} of an 
utterance, followed by a higher-level \textit{premotor representation}, leading to neuromuscular activation 
patterns, and finally to the articulatory-acoustic realization of the utterance (cf.~various hierarchical 
models in the literature).

The cognitive-linguistic representation may be viewed as the activation of a phonological sequence, 
often organized into syllables. For instance, the two-syllabic word /\ipa{kam.flik}/  (nonsense 
word in Indo-European languages) can be decomposed into two syllables, each containing a consonantal 
on- and offset and a vocalic center. The higher-level motor representation already includes an 
estimation of sound durations and of the associated articulator movements.

A successful concept for describing motor representations at this intermediate level is the 
\textbf{gesture score} concept (\citet{Browman_Goldstein_1992, Pouplier_Goldstein_2005, Kroeger_Birkholz_2007, Kroeger_etal_2022, 
Kroeger_2023}). \textit{Gestures} represent goal-directed articulatory actions such as: (i) labial, 
apical, or dorsal closing gestures for the realization of consonantal constrictions, (ii) velopharyngeal 
opening or closing gestures for the production of nasal or non-nasal sounds, (iii) glottal opening 
or closing gestures for voiceless or voiced sounds, and (iv) vocalic gestures defining the overall 
shaping of the vocal tract for specific vowels.

Each gesture defines the temporal interval and movement trajectory of one or more articulators toward 
a specified target, including the holding phase during which the articulators maintain the target 
configuration. The coordination and timing (onset, duration, and overlap) of all gestures within 
an utterance are captured by the \textbf{gesture score}, which defines the 
\textit{premotor representation} of speech (cf. \citet{Kroeger_etal_2020, Kroeger_etal_2022, Kroeger_2023}).

In this paper, we describe a concrete computer implementation of gestures and gesture scores as well 
as the resulting \textit{computation of articulator movements} within the articulatory model DYNARTmo.

\section{The Gesture Approach of DYNARTmo}
\subsection{Qualitative considerations: cognitive-linguistic level}

Our modeling approach, \textbf{DYNARTmo}, simulates six model articulators—\textit{lips, tongue tip, 
tongue dorsum, lower jaw, velum, and glottis}. Its movements are controlled by a set of ten 
continuously scaled control parameters (displacement control) and six discrete parameters (control of 
manner and place of consonantal constrictions, see \citet{Kroeger_2025a}).

\begin{figure}[!htbp]
\centering
\includegraphics[width=0.95\textwidth]{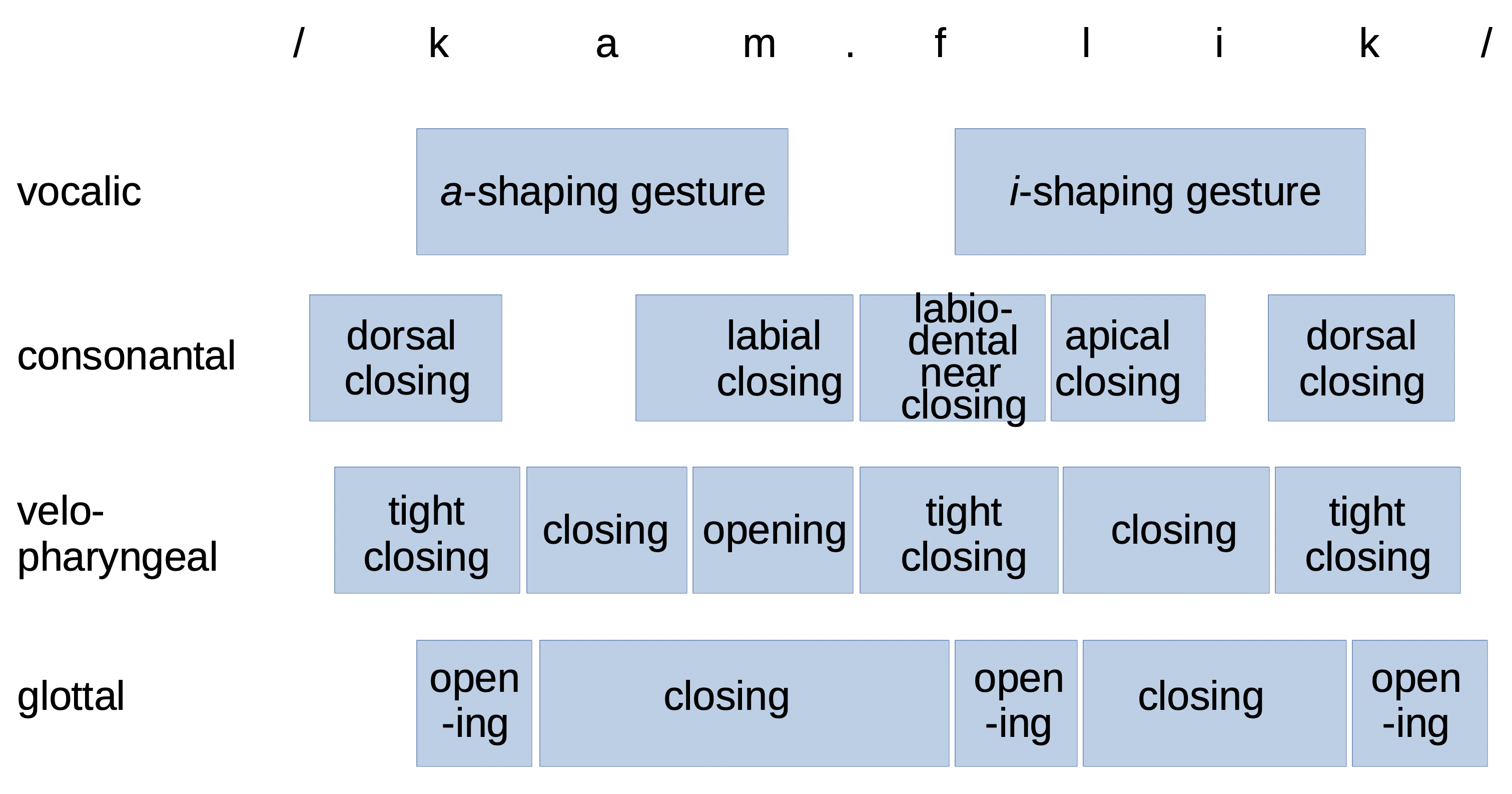}
\caption{Cognitive-linguistic (phonological) representation of gesture score of the two-syllabic 
(nonsense) word “kamflik” /kam.flik/. 
Gestures are arranged in temporal order across four gesture tiers: 
vocalic, consonantal, velopharyngeal, and glottal gesture tier.
Blue rectangles indicate duration of each gesture (gesture activation interval).}
\label{fig:gesture_score}
\end{figure}

A subset of the implemented gesture inventory is outlined in Table~\ref{tab:gestures}. The current 
gesture set represents \textit{Standard German} articulatory gestures. The concept of a gesture 
provides an elegant solution to the long-standing \textit{phonology-phonetics interface problem}. 
On the one hand, phonemes can be represented by sets of gestures (see Table~\ref{tab:phonemes2gestures}), 
so that the discrete gesture score (cf. \citet{Kroeger_etal_2020, Kroeger_etal_2022}) can be interpreted 
as a \textit{phonological or cognitive-linguistic representation} of a syllable, word, or utterance 
(see Figure~\ref{fig:gesture_score}). On the other hand, each gesture specifies a continuous 
articulatory movement pattern and can therefore be quantitatively defined in a 
\textit{phonetic-motoric} sense.

\begin{table}[!htbp]
\centering
\small
\caption{Gestures of Standard German: type, name, main articulators, and 
corresponding sounds in IPA and SAMPA notation (following \citet{Kroeger_Birkholz_2007}).}
\label{tab:gestures}
\begin{tabular}{llll}
\toprule
\textbf{Name} &
\textbf{Main articulators} &
\makecell{\textbf{Sounds}\\\textbf{IPA}} &
\makecell{\textbf{Sounds}\\\textbf{SAMPA}} \\
\midrule
\multicolumn{4}{l}{\textbf{Vocalic gestures:}}\\
vocalic \textit{ii}-shaping  & tongue, lips & /\ipa{i:}/  & /i:/ \\
vocalic \textit{is}-shaping  & tongue, lips & /\ipa{I}/   & /I/ \\
vocalic \textit{uu}-shaping  & tongue, lips & /\ipa{u:}/  & /u:/ \\
vocalic \textit{us}-shaping  & tongue, lips & /\ipa{U}/   & /U/ \\
vocalic \textit{aa}-shaping  & tongue, lips & /\ipa{a:}/  & /a:/ \\
vocalic \textit{ae}-shaping  & tongue, lips & /\ipa{E:}/  & /E:/ \\
vocalic \textit{yy}-shaping  & tongue, lips & /\ipa{y:}/  & /y:/ \\
vocalic \textit{22}-shaping  & tongue, lips & /\ipa{\o:}/ & /2:/ \\
vocalic \textit{schwa}-shaping & tongue, lips & /\ipa{@}/ & /@/ \\
vocalic \textit{a6}-shaping  & tongue, lips & /\ipa{6}/   & /6/ \\
\ldots & \ldots & \ldots & \ldots \\
\midrule
\multicolumn{4}{l}{\textbf{Consonantal gestures:}}\\
labial closing & lips & /\ipa{p, b, m}/ & /p, b, m/ \\
apical closing (raising) & tongue tip & /\ipa{t, d, n}/ & /t, d, n/ \\
dorsal closing (raising) & tongue dorsum & /\ipa{k, g, N}/ & /k, g, N/ \\
labio-dental near closing & lips & /\ipa{f, v}/ & /f, v/\\
dental near closing & tongue tip & /\ipa{T, D}/ & /T, D/\\
alveolar near closing & tongue tip & /\ipa{s, z}/ & /s, z/\\
postalveolar near closing & tongue tip & /\ipa{S, Z}/ & /S, Z/\\
palatal near closing & tongue dorsum & /\ipa{C, J}/ & /C, J/\\
velar/uvular near closing & tongue dorsum & /\textipa{x, G, X}/ & /x, G, X/\\
apical lateral closing & tongue tip & /\ipa{l}/ & /l/ \\
palatal approximant & tongue dorsum & /\ipa{j}/ & /j/\\
labial-velar approximant & tongue dorsum, lips & /\ipa{w}/ & /w/\\
\ldots & \ldots & \ldots & \ldots \\
\midrule
\multicolumn{4}{l}{\textbf{Velopharyngeal gestures:}}\\
\makecell{velopharyngeal closing\\(velum raising)} & velum & non-nasal sounds \\
\makecell{velopharyngeal tight closing\\(velum strong raising)} & velum & obstruents \\
\makecell{velopharyngeal opening\\(velum lowering)} & velum & nasals \\
\ldots & \ldots & \ldots & \ldots \\
\midrule
\multicolumn{4}{l}{\textbf{Glottal gestures:}}\\
glottal closing & glottis & voiced sounds \\
glottal tight closing & glottis & /\ipa{P}/ & /?/ \\
glottal opening & glottis & voiceless sounds \\
\ldots & \ldots & \ldots & \ldots \\
\bottomrule
\end{tabular}
\end{table}

\begin{table}[!htbp]
\centering
\small
\caption{Activated gestures per phoneme or speech sound. Each sound comprises three 
gestures in most cases. One gesture is activated in case of glottal consonants.   
The exact naming of gestures as well as the exact complete set of phonemes is 
language specific.}
\label{tab:phonemes2gestures}
\begin{tabular}{llll}
\toprule
\textbf{Phoneme} & \textbf{\makecell{Tract\\gesture}} & \textbf{\makecell{Velopharyngeal\\gesture}} & \textbf{\makecell{Glottal\\gesture}} \\
\midrule
\multicolumn{4}{l}{\textbf{Vowels:}} \\
/a/ & vocalic \textit{a}-shaping & closing & closing \\
/i/ & vocalic \textit{i}-shaping & closing & closing \\
/u/ & vocalic \textit{u}-shaping & closing & closing \\
/\ipa{@}/ & vocalic \textit{schwa}-shaping & closing & closing \\
\ldots & \ldots & \ldots & \ldots \\
\midrule
\multicolumn{4}{l}{\textbf{Consonants (obstruents, voiceless):}} \\
/p/ & labial closing & tight closing & opening \\
/k/ & dorsal--velar closing & tight closing & opening \\
/s/ & apical--alveolar near closing & tight closing & opening \\
/f/ & labiodental near closing & tight closing & opening \\
\ldots & \ldots & \ldots & \ldots \\
\midrule
\multicolumn{4}{l}{\textbf{Consonants (obstruents, voiced):}} \\
/b/ & labial closing & tight closing & closing \\
/g/ & dorsal--velar closing & tight closing & closing \\
/z/ & apical--alveolar near closing & tight closing & closing \\
/v/ & labiodental near closing & tight closing & closing \\
\ldots & \ldots & \ldots & \ldots \\
\midrule
\multicolumn{4}{l}{\textbf{Consonants (sonorants, voiced):}} \\
/l/ & apical--lateral closing & closing & closing \\
/j/ & dorsal--palatal approx. closing & closing & closing \\
/m/ & labial closing & opening & closing \\
/n/ & apical closing & opening & closing \\
\ldots & \ldots & \ldots & \ldots \\
\midrule
\multicolumn{4}{l}{\textbf{Consonants (glottal):}} \\
/\ipa{P}/ & glottal tight closing & -- & -- \\
/h/ & glottal near opening & -- & -- \\
\ldots & \ldots & \ldots & \ldots \\
\bottomrule
\end{tabular}
\end{table}

\subsection{Quantitative considerations: phonetic-motoric level}

\begin{figure}[!htbp]
\centering
\includegraphics[width=1.0\textwidth]{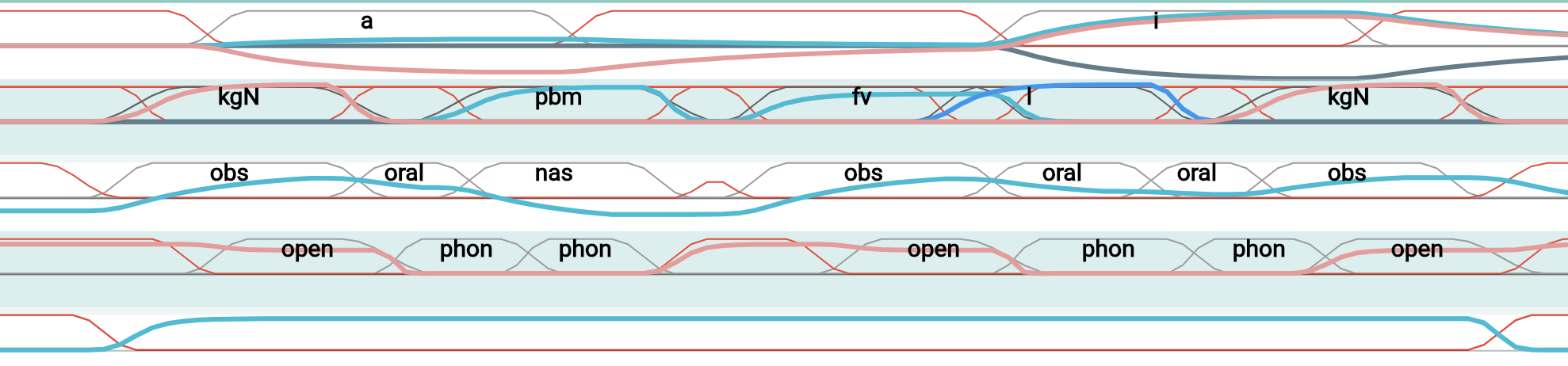}
\caption{Phonetic-motoric representation of the gesture score and control parameter trajectories for 
main articulators for the same nonsense word “kamflik” as 
displayed in Figure~\ref{fig:gesture_score}. 
Gesture activation levels (thin gray-black lines) and activation levels of neutral gestures 
(thin red lines), as well as control parameter trajectories for the main articulators (green-blue, 
red, and gray-black thick lines), are shown. 
Gestures and control parameter trajectories are arranged according to gesture tiers: 
vocalic, consonantal, velopharyngeal, glottal, and pulmonary gesture tier.
\textbf{Vocalic tier} (white background): a- and i-shaping gesture:  this tier indicates vocalic height as red, 
vocalic position (fronting) as green-blue, and lip rounding (spreading is indicated by negative values) 
as gray-black line;
\textbf{consonantal tier} (light blue-green background): dorsal closing (= kgN), labial closing (= pbm), 
labio-dental near closing (= fv), apical lateral closing (= l), and again dorsal closing gesture; 
this tier indicates labial closing as green-blue line, apical closing as dark blue, and 
dorsal closing as red thick line; values on the consonantal tier are positive indicating 
degree of consonantal constriction; 
\textbf{velopharyngeal tier} (white background): velopharyngeal tight closing (= obstruent), 
closing (= oral), openig (= nasal) and so forth gestures; the tier indicates velopharyngeal 
opening or closing up to tight closing as thick green-blue line (opening as negative values 
indicating velum position and closure up to tight closure as positive values); 
\textbf{glottal tier} (light blue-green background): opening (= open), closing (= phonation) and so forth gestures;
the tier indicates degree of glottal opening (glottal abduction) as thick red line; 
\textbf{pulmonary tier} (white background): one pulmonary gesture (green-blue thick line) for producing 
constant subglottal pressure during speaking.}
\label{fig:gesture_dynamics}
\end{figure}

The relevant phonetic-motoric parameters include: (i) the \textbf{duration} of a gesture (its 
onset and offset times), (ii) the \textbf{main articulator} of a gesture, (iii) the 
\textbf{main articulator lateral shape}, (vi) the \textbf{spatial target} of a gesture, 
and (v) the \textbf{rapidity} (velocity) with which the main articulator approaches the target region 
after gesture activation (see Table~\ref{tab:gesture_params}). The timing and coordination (sequential as well as 
temporally overlapping) of gestures are explicitly defined in the gesture score 
(see Table~\ref{tab:gesture_params}, and see Figure~\ref{fig:gesture_dynamics}).

\vspace{0.5em}

{\small
Phonetic-motoric gesture parameters are exemplified for vocalic and consonantal gestures in 
Table~\ref{tab:gesture_params}. 
\textbf{Lateral shape} indicates manner of articulation (mid = central groove 
in tongue as appearing in fricatives; lateral = lowering of lateral parts of tongue as 
appearing in lateral sounds; normal = no groove or lateral lowering which allows vocalic 
tongue and lips shaping or which allows full vocal tract closure in case of plosives and nasals; 
see also \citet{Kroeger_2025b}). 
\textbf{Target values}: 
Vocalic gestures --- Vocalic height (\textit{hei}), vocalic position or fronting 
(\textit{pos}), lips rounding (\textit{rou}) vary between $-1$ and $1$ 
(see \citet{Kroeger_2025a}); zero and negative lip rounding indicates spreading. 
Consonantal gestures --- Closure values (\textit{clo}) above 1 indicate tight closure. 
Velopharyngeal gestures --- Negative closure values (\textit{clo}) indicate velum lowering, 
positive values indicate degree of tightness of velopharyngeal closure. 
Glottal gestures --- Positive opening values (\textit{opg}) indicate degree of glottal 
opening, negative values indicate tight closure. 
\textbf{Duration}: only mean values. Concrete gesture duration is calculated at the level 
of integrating gestures into gesture score. 
\textbf{Rapidity} (velocity of gesture execution) equals the reciprocal pull factor (see text). 
All values are estimated from modeling considerations and optimized to fit articulatory 
demands for generating valid sound durations.}

\begin{table}[H]
\centering
\small
\caption{Phonetic-motoric gesture parameters for vocalic and 
consonantal gestures.}
\label{tab:gesture_params}
\begin{tabular}{llllll}
\toprule
\textbf{\makecell{Gesture\\name}} &
\textbf{\makecell{Main\\articulators}} &
\textbf{\makecell{Lateral\\shape}} &
\textbf{\makecell{Target\\value}} &
\textbf{\makecell{Duration\\(ms)}} &
\textbf{Rapidity} \\
\midrule
\multicolumn{6}{l}{\textbf{Vocalic gestures:}} \\
\makecell[l]{vocalic\\a-shaping} & \makecell[l]{tongue\\lips} & normal & 
\makecell[l]{hei = $-0.80$\\pos = $0.20$\\rou = $0.00$} &  150 & 22.5 \\

\makecell[l]{vocalic\\i-shaping} & \makecell[l]{tongue\\lips} & normal & 
\makecell[l]{hei = $0.90$\\pos = $1.00$\\rou = $-1.00$} & 150 & 22.5 \\

\makecell[l]{vocalic\\u-shaping} & \makecell[l]{tongue\\lips} & normal & 
\makecell[l]{hei = $0.90$\\pos = $-1.00$\\rou = $1.00$ } & 150 & 22.5 \\

\makecell[l]{vocalic\\schwa-shaping} & \makecell[l]{tongue\\lips} & normal & 
\makecell[l]{hei = $-0.80$\\pos = $0.20$\\rou = $0.00$} & 100 & 22.5 \\

\midrule
\multicolumn{6}{l}{\textbf{Consonantal gestures:}} \\
\makecell[l]{labial\\closing} & lips & normal & clo = 1.07 & 100 & 30.0 \\
\makecell[l]{labio--dental\\near closing} & lips & to teeth & clo = 0.80 & 100 & 30.0 \\
\makecell[l]{apical\\closing} & tongue tip & normal & clo = 1.07 & 100 & 30.0 \\
\makecell[l]{lateral apical\\closing} & tongue tip & lateral & clo = 1.07 & 100 & 30.0 \\
\makecell[l]{dorsal\\closing} & tongue dorsum & normal & clo = 1.07 & 100 & 30.0 \\
\midrule
\multicolumn{6}{l}{\textbf{Velopharyngeal gestures:}} \\
\makecell[l]{velopharyngeal\\opening} & velum & normal & clo = $-0.5$ & 100 & 22.5 \\
\makecell[l]{velopharyngeal\\closing} & velum & normal & clo = $0.0$ & 100 & 22.5 \\
\makecell[l]{velopharyngeal\\tight closing} & velum & normal & clo = $0.5$ & 100 & 22.5 \\
\midrule
\multicolumn{6}{l}{\textbf{Glottal gestures:}} \\
\makecell[l]{glottal\\opening} & glottis & normal & opg = $0.01$ & 100 & 90.0 \\
\makecell[l]{glottal\\closing} & glottis & normal & opg = $0.5$ & 100 & 90.0 \\
\makecell[l]{glottal tight\\closing} & glottis & normal & opg = $0.5$ & 100 & 90.0 \\
\bottomrule
\end{tabular}
\end{table}

\subsection{Mathematical model for gesture trajectories}

Each gesture is defined as a temporally localized activation pattern 
that drives a subset of articulatory control parameters (main articulator plus, see 
\citet{Kroeger_2025a}).  
A gesture instance $g$ is characterized by the tuple
\[
(t_s, t_e, \tau_{\text{on}}, \tau_{\text{off}}, \mathbf{T}_g, p_g)
\]
where $t_s$ and $t_e$ denote the onset and offset times (in~ms), 
$\tau_{\text{on}}$ and $\tau_{\text{off}}$ specify the duration of the rising and 
falling flanks of the activation, $\mathbf{T}_g = \{T_{g,1}, T_{g,2}, \dots\}$ 
represents the gesture's target values for the relevant control parameters, 
and $p_g$ is a weighting factor (``pull'') describing the relative strength 
of the gesture.

\paragraph{Activation function.}
The temporal activation of gesture $g$ is modeled by a smooth cosine-shaped
function $a_g(t)$ following the formulation of \citet{Saltzman_Munhall_1989} 
and \citet{Nam_etal_2006} within the Task-Dynamics framework:
\begin{align*}
a_g(t) = 
\begin{cases}
0, & t < t_s,\\[4pt]
\dfrac{1}{2}\,\left(1 - \cos\,\left(\pi \dfrac{t - t_s}{\tau_{\mathrm{on}}}\right)\,\right),
 & t_s \le t < t_s + \tau_{\mathrm{on}},\\[8pt]
1, & t_s + \tau_{\mathrm{on}} \le t < t_e - \tau_{\mathrm{off}},\\[4pt]
\dfrac{1}{2}\,\left(1 + \cos\,\left(\pi \dfrac{t - (t_e - \tau_{\mathrm{off}})}{\tau_{\mathrm{off}}}\right)\,\right),
 & t_e - \tau_{\mathrm{off}} \le t < t_e,\\[4pt]
0, & t \ge t_e.
\end{cases}
\end{align*}

\begin{figure}[!htbp]
\centering
\includegraphics[width=1.0\textwidth]{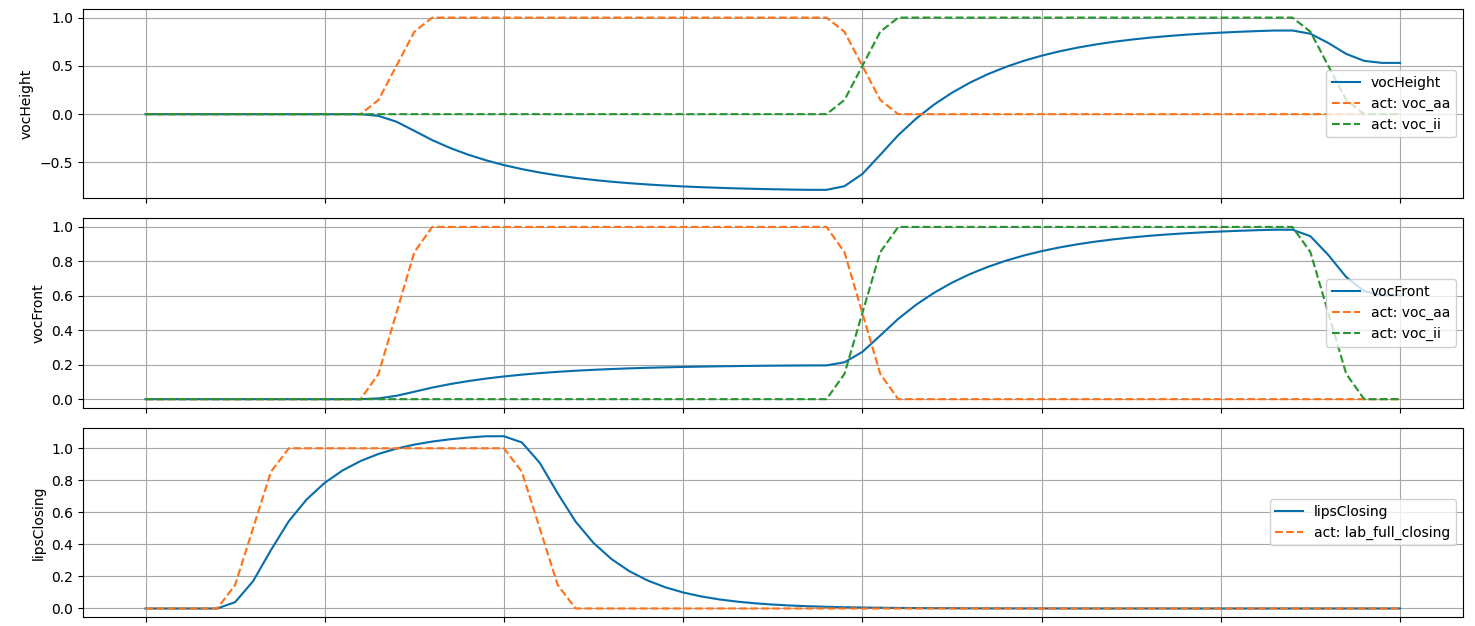}
\caption{Gesture activation functions (dashed lines) and resulting control parameter trajectories (solid lines) 
for three control parameters of our articulatory model (vocalic height, vocalic position or fronting, and labial 
closing) are given for /\ipa{pa:i:}/. Three gestures are shown: two vocalic gestures, 
i.e., an a-shaping followed by an i-shaping gesture and one consonantal gesture, i.e., a labial closing gesture.} 
\label{fig:gesture_python}
\end{figure}

This function rises and falls smoothly with continuous first derivative ($C^1$), 
producing the cosine‑shaped temporal activation functions as shown in Figure~\ref{fig:gesture_dynamics}
and in Fig~\ref{fig:gesture_python}.

\paragraph{Calculation of articulator displacement.}
The instantaneous displacement $D_{g,P}(t)$ of the articulator governed by gesture~$g$ 
for a given control parameter $P$ is defined as the product of the gesture’s temporal activation $a_g(t)$ 
and its associated target value $T_{g,P}$:
\[
D_{g,P}(t) = a_g(t)\, T_{g,P}.
\]
This function describes a smooth, time‑varying trajectory for the control parameter $P$ 
under the influence of gesture $g$ alone.

\paragraph{Weighted blending across overlapping gestures.}
When multiple gestures simultaneously affect the same control parameter $P$, 
their individual displacements $D_{g,P}(t)$ are combined through a normalized weighted sum, 
using the pull parameter $p_g$ as a weighting factor:
\[
P(t) = 
\frac{\displaystyle\sum_{g \in G_t} p_g\, D_{g,P}(t)}
     {\displaystyle\sum_{g \in G_t} p_g\, a_g(t)} 
= 
\frac{\displaystyle\sum_{g \in G_t} p_g\, a_g(t)\, T_{g,P}}
     {\displaystyle\sum_{g \in G_t} p_g\, a_g(t)} ,
\]
where $G_t$ is the set of all gestures that are active for parameter $P$ at time $t$.

The degree of influence of individual gestures is governed by their pull parameter $p_g$, 
which modulates their relative weight in the blending process.

If no gesture is active for a parameter at a given time point, its trajectory defaults to zero 
or to a speaker‑specific neutral value (see below).

\paragraph{Discrete‑time implementation.}
The trajectories are computed at discrete time steps $\Delta t$ (e.g.\ 5\,ms).  
For each sample $t_i = i\,\Delta t$:
\begin{enumerate}
  \item Identify all gestures $g$ for which $t_s \le t_i < t_e$,
  \item evaluate the activation $a_g(t_i)$ using the equations above,
  \item compute the displacement $D_{g,P}(t_i) = a_g(t_i)\, T_{g,P}$ for each relevant parameter $P$,
  \item and calculate the blended parameter value $P(t_i)$ via the weighted sum over all active gestures.
\end{enumerate}
This procedure yields smooth, time-resolved articulatory trajectories $P(t)$ 
for all control parameters involved in a given syllable or utterance.

\vspace{1.5em}

This mathematical formulation provides continuous transitions between overlapping gestures, 
thus offering a quantitative account of articulation and coarticulation phenomena.

\section{Calculating gesture scores by rule}

The activation of the quantitative gesture score takes place at premotor level within the cortex 
(see \citet{Kroeger_etal_2022}). It can be assumed that the gesture scores of most frequent syllables 
of a language are stored at this level. This leads to a learning process for extracting basic rules for
language specific syllable construction (i.e. gesture score construction of less frequent syllables in 
a specific mother tongue; see dual route model: \citet{Whiteside_Varley_1998}; \citet{Samlowski_2014}; 
\citet{Miller_2020}). 

We tried to implement a rough basic rule system for calculating the on- and offset times for 
all gestures activated during syllable production, but the resulting rule system is still erroneous because
it seems nearly impossible to generate gesture scores and thus to generate articulatory trajectories from 
a phonological specification of a word or utterance directly by rule, i.e. without accessing a syllable 
lexicon. Thus, a concept of a mental syllabary following the ideas of \citet{Levelt_Wheeldon_1994}, 
\citet{Cholin_2008}, and \citet{Pan_Zhang_2023} should be integrated at this level in order to enable a 
precise language specific quantification of gesture scores for syllables, words and short utterances.

The onset and offset values generated for the (nonsensce) word "Kamflik" by our current version of the DYNARTmo 
rule-system are given in Table~\ref{tab:gesture_timing}. The appropriate display of gestural score is given in 
Fig~\ref{fig:gesture_dynamics}. In those time intervals in which no linguistic relevant gesture is active, the 
model activates a neutral gesture. The set of gestures representing the neutral state comprises a schwa-like 
positioning of the model articulators accompanied by a glottal opening (for breathing) and further accompanied 
by a velopharyngeal opening for allowing to inhalate and exhalate via mouth as well as via nose during silent 
phases which appear before and after the production of an utterance.

\begin{table}[!htbp]
\centering
\small
\caption{Onset and offset times of gesture activation for all gestures of the nonsense word “kamflik” /kam.flik/. 
Directly adjacent gestures of the same type within the velopharyngeal or glottal tiers 
(e.g., two consecutive velopharyngeal closing gestures or two consecutive glottal closing gestures) 
may merge into one gesture; see Figure~\ref{fig:gesture_dynamics}.}
\label{tab:gesture_timing}
\begin{tabular}{lllll}
\toprule
\textbf{Sound} & \textbf{Gesture name} & \textbf{\makecell{Onset\\(ms)}} & \textbf{\makecell{Offset\\(ms)}} & \textbf{\makecell{Duration\\(ms)}} \\
\midrule
/k/ & dorsal closing & 50 & 150 & 100 \\
& velopharyngeal tight closing & 50 & 145 & 95 \\
& glottal opening & 80 & 160 & 80 \\
\addlinespace[2pt]
/a/ & vocalic \textit{a}-shaping & 90 & 200 & 110 \\
& velopharyngeal closing & 125 & 180 & 55 \\
& glottal opening & 140 & 200 & 60 \\
\addlinespace[2pt]
/m/ & labial closing & 150 & 250 & 100 \\
& velopharyngeal opening & 160 & 240 & 80 \\
& glottal closing & 180 & 235 & 55 \\
\addlinespace[2pt]
/f/ & labiodental near closing & 250 & 350 & 100 \\
& velopharyngeal tight closing & 250 & 345 & 95 \\
& glottal opening & 280 & 355 & 75 \\
\addlinespace[2pt]
/l/ & apical lateral closing & 310 & 400 & 90 \\
& velopharyngeal closing & 325 & 395 & 70 \\
& glottal closing & 355 & 410 & 75 \\
\addlinespace[2pt]
/i/ & vocalic \textit{i}-shaping & 340 & 450 & 110 \\
& velopharyngeal closing & 375 & 430 & 55 \\
& glottal closing & 390 & 450 & 60 \\
\addlinespace[2pt]
/k/ & dorsal closing & 400 & 500 & 100 \\
& velopharyngeal tight closing & 410 & 490 & 80 \\
& glottal opening & 430 & 510 & 80 \\
\bottomrule
\end{tabular}
\end{table}

\section{Integrating the control concept into DYNARTmo}
The Speech Articulation Trainer Web App (\url{https://speech-articulation-trainer.web.app/}) 
allows (i) the animation of vowels (vow button), (ii) the animation of consonants 
(cons button), both displaying still images of the chosen speech sound, and allows 
(iii) the prompting of syllables (syll button) and (iv) the generation of video animations 
of the syllable sequence (move button). The generation of still images (sagittal, glottal 
and palatal view) are explained in detail in \citet{Kroeger_2025a} and \citet{Kroeger_2025b}. 

Currently the app allows the generation of movement animations of up to three syllables with 
up to 2 consonants in onset and in offset and up to three vocalic gestures in center of 
each syllable. In Standard German a succession of two vocalic gestures is needed in order to 
simulate diphthongs like /\ipa{aI}/, /\ipa{aU}/ or /\ipa{OI}/ while a succession of three 
vocalic gestures is needed if a vocalized /r/ is following a diphthong like in the word 
"Feier" /\ipa{faI6}/ (party). The maximum of two consonants in on- and offset allows the 
generation of German words like "spielt" /\ipa{Spi:lt}/ (is playing) or "Klatsch" /\ipa{klatS}/ 
(gossip) and so on.    

\section{Discussion and Conclusion}

The DYNARTmo model provides a computationally explicit implementation of the gesture-based approach 
to speech motor control (see also \citet{Kroeger_2025a, Kroeger_2025b}). By linking phonological representations (gesture scores) with quantitative 
articulatory trajectories, DYNARTmo bridges the gap between abstract linguistic planning and the 
biomechanical realization of speech. Future work will integrate the model with neural-level control 
architectures and extend the gesture inventory to cover multiple languages.

\section{Supplementary Material}

The gesture and control parameter (or movement) trajectory generator as part of the articulatory model 
is available for download. This module is written in Python and is located in the folder 
\texttt{py\_code/gestures\_and\_trajectories.zip}. After unzipping, the program can be 
executed in any Python interpreter by running the file \texttt{demo\_pai\_lex.py} and that
program will generate the data for Figure~\ref{fig:gesture_python} and the figure itself.

\bibliographystyle{apalike}
\bibliography{references}

\end{document}